\documentclass[journal]{IEEEtran}
\usepackage{amsmath}
\usepackage{mathrsfs}
\usepackage{amsfonts}
\usepackage[numbers,sort&compress]{natbib}
\usepackage{amssymb}
\usepackage{bm}
\usepackage{enumerate}
\usepackage[pdftex]{graphicx}
\usepackage{lineno,hyperref}
\emergencystretch=\maxdimen
\ifCLASSINFOpdf
\else
\fi

\begin{document}
%
\title{Local Activity-tuned Image Filtering for \\Noise Removal and Image Smoothing}
%
%
%

\author{Lijun~Zhao,
        Jie~Liang,~\IEEEmembership{Senior~Member,~IEEE,}
        Huihui~Bai,~\IEEEmembership{Member,~IEEE,}
        Lili Meng,
        Anhong~Wang,~\IEEEmembership{Member,~IEEE,}
        and~Yao~Zhao,~\IEEEmembership{Senior~Member,~IEEE}
\thanks{L.  Zhao, H. Bai, Y. Zhao are with Institute Information Science, Beijing Jiaotong University, Beijing, 100044, P. R. China, e-mail: {15112084, hhbai, yzhao}@bjtu.edu.cn.}
\thanks{J. Liang is with School of Engineering Science, Simon Fraser University, ASB 9843, 8888 University Drive, Burnaby, BC, V5A 1S6, Canada, e-mail:jliang@sfu.ca}
\thanks{L. Meng is with School of Information Science and Engineering, Shandong Normal University, Jinan, 730050, P. R. China, e-mail: mengll83@hotmail.com}
\thanks{A. Wang is with Institute of Digital Media \& Communication, Taiyuan University of Science and Technology, Taiyuan, 030024, P. R. China, e-mail:wah\_ty@163.com}}

\maketitle

\begin{abstract}
In this paper, two local activity-tuned filtering frameworks are proposed for noise removal and image smoothing, where the local activity measurement is given by the clipped and normalized local variance or standard deviation. The first framework is a modified anisotropic diffusion for noise removal of piece-wise smooth image. The second framework is a local activity-tuned Relative Total Variation (LAT-RTV) method for image smoothing. Both frameworks employ the division of gradient and the local activity measurement to achieve noise removal. In addition, to better capture local information, the proposed LAT-RTV uses the product of gradient and local activity measurement to boost the performance of image smoothing. Experimental results are presented to demonstrate the efficiency of the proposed methods on various applications, including depth image filtering, clip-art compression artifact removal, image smoothing, and image denoising.
\end{abstract}


%
\IEEEpeerreviewmaketitle

\section{Introduction}
%
%
%
%
\IEEEPARstart{I}{mage} filtering is an effective way to improve the performance of many applications, such as rain removal \cite{R31D}, stereo matching \cite{R6D,R7J,R32S}, edge detection, and image editing \cite{R17L,R18L,R21L,R22K,R23E,R24L,R25Q,R26B}. Since different types of images have different characteristics and different applications have different requirements, the filtering algorithms should be designed for each case properly. For example, depth images are mainly determined by the scene's geometry, and typically have smooth regions with sharp boundaries. The boundaries should be preserved with high quality, as it will affect the quality of depth image-based rendering (DIBR), view synthesis, and 3D video coding's efficiency \cite{R35C,R36C,C45}. On the other hand, for natural images, if we want to remove the noise, we need to preserve both the image's structure and textural information. If we want to apply image smoothing, we should remove the detailed textures, but keep the major structural information.

Bilateral filter is an important image filtering technique \cite{R27K}, which can remove image noises and preserve sharp boundaries. A fast bilateral filtering is developed in \cite{R28K}. In \cite{R29K}, an optimally weighted bilateral filter is proposed, whose performance is competitive to the non-local means filter \cite{R30A}. With self-learning based image decomposition for single image denoising, the undesirable patterns are automatically determined by the derived image components directly from the input image \cite{C46}. Anisotropic diffusion is another well-known image denoising algorithm \cite{R1P}. The relationship between anisotropic diffusion and robust statistics is analyzed in \cite{R2M}. In \cite{R5J}, a new class of fractional-order anisotropic diffusion equations is introduced for noise removal, where the discrete Fourier transform is used and an iterative scheme in the frequency domain is also given. A noise removal filter is built by an image activity detector based on the density of connected components \cite{S43}. Latter, a set of textures and images is analyzed to determine the best measure of image activity and it has showed that image activity measure has powerful ability to capture the activities and differentiating between various images \cite{W44}. To preserve edges and fine details while effectively removing noise, both local gradient and variance are incorporated into the diffusion model \cite{R3S,R4S}.  In \cite{R6D,R7J,R8S,R9D,R10L}, anisotropic diffusion is applied to 3D image processing fields. In \cite{R11NN}, anisotropic diffusion is utilized as a preprocessing of DIBR to improve its quality.

To remove severe artifacts in the compressed depth images, many methods have been explored to filter depth images so as to improve the quality of the synthesized virtual images. In \cite{R12S}, a trilateral filtering method is treated as an in-loop filter to prevent depth coding artifacts. This method employs spatial domain filter, depth range domain filter, and color range domain filter. In \cite{R13X}, an adaptive depth truncation filter (ADTF) is presented to restore the sharp object boundaries of depth images. In \cite{R14L}, a candidate-value-based depth boundary filtering (CVBF) is developed by selecting an appropriate candidate value to replace each unreliable pixel according to both spatial correlation and statistical characteristics. Recently, a two-stage filtering (TSF) scheme is proposed in \cite{R33L}, using binary segmentation-based depth filtering and Markov Random Field (MRF). These methods greatly reduce the coding artifacts in the synthesized virtual images, but they often change depth images too much.

Image smoothing is another important technique for many applications. Generally, image smoothing can be classified into two classes: weighted filtering methods and optimization-based methods. Weighted filtering is usually achieved by a weighting method within a window. For example, the guided image filter in \cite{R22K} is a fast and non-approximate linear time algorithm. It has nothing to do with the kernel size and the intensity range. Another efficient method is the rolling guidance filtering \cite{R25Q}, which is a fast iterative method based on bilateral filtering. For real-time tasks, a high-quality edge preserving filtering is proposed in \cite{R23E}.

Different from these weighted filtering methods, optimization-based smoothing methods always face a non-convex yet complex problem. In \cite{R26B}, both the static guidance and dynamic guidance are jointly leveraged to achieve robust guided image filtering, which is formulated as a nonconvex optimization problem. In \cite{R21L}, a multi-scale image decomposition method is presented with weighted least square optimization framework to form edge-preserving smoothing operator. In \cite{R18L}, an L0 gradient minimization optimization framework is proposed, which globally controls how many non-zero gradients are kept in the filtered image. By taking advantage of the statistic diversity of gradient information between texture patches and structure patches, the Relative Total Variation (RTV) framework is proposed in \cite{R17L}. In this method, the inherent variation and total variation are combined together to discriminate the structure from texture, and an optimization problem is formulated to extract the main structure of the image. Later, another efficient image smoothing approach is proposed based on region covariance \cite{R24L}. Although these methods achieve excellent performances for structure-preserving smoothing, there are still some problems, such as inefficient texture removal and severe edge blurring after smoothing.

In this paper, the clipped and normalized local variance or standard deviation (std) is used as the local activity measurement. In addition, both image gradient and local activity are exploited for image smoothing and denoising. In particular, we show that the product of the gradient and the clipped local activity can better seize the change of the image around a pixel in the presence of noise, while the ratio between the gradient and the clipped local activity could locate the noises in the image and facilitate denoising. In our first framework, we develop a robust local activity-tuned anisotropic diffusion framework and apply it for compression artifact removal of piece-wise smooth images such as depth images and clip-art images.

Our second framework uses a local activity-tuned relative total variation, which includes two schemes. The first scheme is a local activity-tuned RTV for image smoothing and image representation in different scale-spaces, where the RTV is divided by the clipped local activity, which emphasizes the contour information of the image. The second local activity-tuned RTV scheme is designed to remove additive white Gaussian noise, which uses the ratio between the gradient and local activity. This can identify the location of the noise. The performances are demonstrated by experimental results.

The rest of this paper is organized as follows. In Section II, a robust local activity-tuned anisotropic diffusion scheme is described. In Sec. III, a local activity-tuned relative total variation framework is introduced. Experimental results are presented in Section IV, followed by the conclusion in Section V.
\section{Local activity-tuned anisotropic diffusion}
\subsection{Perona-Malik anisotropic diffusion}
Anisotropic diffusion is an image denoising technique based on the heat equation, which was originally used to describe the change of temperature in a given region over time. In image processing, it can be used to model the change of pixel values during denoising iterations. The heat equation is given by
\begin{equation}
\frac{\partial I}{\partial t}=\nabla \cdot (\nabla I),
\end{equation}
where $\nabla I$ is the gradient of an image $I$ and $\nabla \cdot (\nabla I)$ denotes the divergence of gradient $\nabla I$, i.e., the Laplacian operator of $I$. Therefore, diffusion happens when the divergence is nonzero. This equation has the same diffusion strength in every direction, therefore it is called isotropic diffusion, which inevitably leads to blur.

Contrary to isotropic diffusion, anisotropic diffusion proposed by Perona-Malik regularizes the images to preserve significant edges \cite{R1P}. The anisotropic diffusion model can be written as
\begin{equation}
\frac{\partial I}{\partial t}=\nabla \cdot (c(||\nabla I||) \nabla I),
\end{equation}
where $c(||\nabla I||)$ is an edge-stop function, such that no diffusion happens across the edges in the image. In \cite{R1P}, two gradient-based edge-stop functions are suggested, i.e.,
\begin{equation}
c(||\nabla I||)=exp(-(\frac{||\nabla I||}{\rho})^2)
\end{equation}

\begin{equation}
c(||\nabla I||)=\frac{1}{1+(\frac{||\nabla I||}{\rho})^2},
\end{equation}
where $\rho$ is a parameter to control the strength of $c(||\nabla I||)$.

The discrete form of the anisotropic diffusion equation can be written as
\begin{equation}
I^{t+1}_{i}=I^{t}_{i}+\lambda \sum_{j \in {\mathbb{N}}_i} c(||\nabla I^t_{ij}||)\nabla I^t_{ij},
\end{equation}
where the parameter $\lambda$ adjusts the convergence speed, $t$ is the iteration number, and $\nabla I^t_{ij}$ denotes the gradient between pixel $I_i$ and pixel $I_j$ in the neighboring $\mathbb{N}_i$ around pixel $I_i$.

\subsection{Modified anisotropic diffusion}
In \cite{R15Y}, the local intensity variance is utilized to adapt the diffusion function:
\begin{equation}
\begin{split}
I^{t+1}_{i}=I^{t}_{i}+\lambda \sum_{j \in \mathbb{N}_i} exp(-(\frac{||\nabla I^t_{ij}||}{k_{ij}})^2)\nabla I^t_{ij} \\k_{ij}=k_{max}-V_{ij} \frac{k_{max}-k_{min}}{max(V)},
\end{split}
\end{equation}
where $k_{ij}$ is the diffusion parameter, $V_{ij}$ is the local gray-scale variance around pixel $I_i^0$ in the initial image, and $max⁡(V)$ is the maximal value of the variance. $k_{max}$ and $k_{min}$ are pre-defined maximal and minimal of $k_{ij}$. This technique could remove noises and irrelevant details while preserving sharper boundaries. However, it only uses the variance of the initial image. This is not optimal, because the initial image’s variance cannot catch up with the updated diffused image's information.

Different from \cite{R15Y}, another anisotropic diffusion model is suggested in \cite{R3S}\cite{R4S},
\begin{equation}
\begin{split}
I^{t+1}_{i}=I^{t}_{i}+\lambda \sum_{j \in \mathbb{N}_i} exp(-\frac{||\nabla I^t_{ij}|| k^2_{i,t}}{\rho})^2\nabla I^t_{ij} \\k^2_{i,t}=1+\frac{\sigma^2_{i,t}-min(\sigma^2_t)}{max(\sigma^2_t)-min(\sigma^2_t)}\cdot 254,
\end{split}
\end{equation}
where $max(\sigma^2_t)$ and $min(\sigma^2_t)$ are the maximal and minimal gray-level variance of the diffused image at the t-th iteration, and $\sigma_{i,t}^2$ is the gray-level variance of the i-th pixel. This method incorporates both local gradient and gray-scale variance to preserve edges and fine details while effectively removing noise. Note that Eq. (6) uses the division or ratio of the gradient and the variance, whereas Eq. (7) uses their product.

\subsection{Local activity-tuned anisotropic diffusion}
In general, depth images are characterized by smooth regions with sharp edges. However, after compression, the edges usually suffer from various compression artifacts, which will affect the quality of view synthesis \cite{R40S}. In this paper, we apply the modified anisotropic diffusion to mitigate the coding artifacts of depth images.
We propose a local activity-tuned anisotropic diffusion (LAT-AD) method, which can be written as
\begin{equation}
\frac{\partial I}{\partial t}=\nabla \cdot (c(||\nabla I||,K) \nabla I),
\end{equation}
where $K$ is obtained from the local activity of the image $I$.

Similar to Eq. (6), the discrete version becomes
\begin{equation}
I^{t+1}_{i}=I^{t}_{i}+\lambda \sum_{j \in \mathbb{N}_i} c(||\nabla I^t_{ij}||, K^t_{i})\nabla I^t_{ij},
\end{equation}
where $I^0_i=I_i$ in the first iteration, $K_i^t$ is a clipped and normalized local activity, which will be defined later. Motivated by \cite{R15Y}, we define two new edge-stop functions as follows:
\begin{equation}
c(||\nabla I^t_{ij}||,K^t_j)=exp(-(\frac{||\nabla I^t_{ij}||}{\rho_1 K^t_i})^2)
\end{equation}

\begin{equation}
c(||\nabla I^t_{ij}||,K^t_j)=exp(-(\frac{||\nabla I^t_{ij}||^2}{(\rho_2)^2 K^t_i})),
\end{equation}
where $\rho_1$ and $\rho_2$ are diffusion parameters. Note that $K_i^t$ is squared in Eq. (10), but not in Eq. (11).

Similar to Eq. (6), the ratio of the gradient and local activity is used, which can capture where the coding artifacts exist in the compressed depth image. Moreover, the diffusion parameter is adaptively tuned according to the ratio, such that larger diffusion parameters are assigned to more severely distorted pixels.  Therefore, pixels with larger local activity would receive more diffusion from neighboring pixels than pixels with smaller activity under the control of gradient. This will remove noisy pixels and prevent blurry regions from being heavily diffused.

We next describe how to calculate the clipped and normalized local activity measurement $K_i^t$. First, we calculate the local mean $\bar{I}_i^t$ and standard variation $v_i^t$ of the 8-connected neighborhood around each pixel.
\begin{equation}
\bar{I}_i^t=\frac{1}{9}(I^t_i+\sum_{j \in \mathbb{N}_i} I^t_j)
\end{equation}

\begin{equation}
v^t_i=[\frac{1}{9}((I_i^t-\bar{I}_i^t)^2+\sum_{j \in \mathbb{N}_i}(I^t_j-\bar{I}_i^t)^2)]^{\frac{1}{2}}
\end{equation}

Next, a clipped version of $v_i^t$ is obtained, denoted as $V_i^t$
\begin{equation}
  V_i^t =
  \begin{cases}
     \frac{1}{2}, &\text{if $0 \leqslant v_i^t < \frac{1}{2}$}\\
	 v_i^t, &\text{if $\frac{1}{2} \leqslant v_i^t < h$}\\
     h, &\text{if $h \leqslant v_i^t$},
  \end{cases}
\end{equation}
where $h$ is a pre-defined parameter.

After that, $V_i^t$ is normalized by $max(V^t)$ in Eq. (15), which is the maximal value across the image.
\begin{equation}
\bar{V}_i^t=V_i^t/max(V^t)
\end{equation}

Finally, to make the iteration more stable,  $K_i^t$ is updated from $\bar{V}_i^t$ for every $l$ iterations.
\begin{equation}
  K_i^t =
  \begin{cases}
    \bar{V}_i^t, &\text{if $mod(t,l)=0$}\\
	\bar{V}_i^{t- mod(t,l)}, &\text{if $mod(t,l)\neq 0$}
  \end{cases}
\end{equation}
where  $mod$ denotes the modulo operator. Let $m$ be the maximal number of iterations. The updating interval $l$ is chosen as $l\in[1,m]$.

In the following, the fixed local activity-tuned anisotropic diffusion using Eq. (10) as edge-stop function is denoted as FLAT-AD, the time-updated local activity-tuned anisotropic diffusion with Eq. (10) is denoted as TLAT-AD, and periodically local activity-tuned anisotropic diffusion based on edge-stop function of Eq. (10) is denoted as PLAT-AD. Moreover, when Eq. (11) is used, the three other methods are denoted as FLAT-AD (I), TLAT-AD (I), and PLAT-AD (I) respectively.

When $l$ is set to be 1, it becomes TLAT-AD. If $l$ is larger than 1, but small than $m$, it reduces to PLAT-AD. However, if $l$ is set to be $m$, it becomes FLAT-AD.

Since the differences between neighboring pixel's variance are often relatively greater than the differences of the corresponding standard deviation, when the 8-connected activity is larger than $\frac{1}{2}$, we use the standard variation instead of variance in this paper. To see this, Let $v_a$ and $v_b$ denote two standard deviations and we assume that $v_a \geq$ $\frac{1}{2}$, $v_b\geq$ $\frac{1}{2}$, and $v_a>$ $v_b$. We look at the difference $(v_a-v_b )-(v_a^2-v_b^2)=$ $(v_a-v_b )[1-(v_a+v_b )]$. Based above assumptions, $1-(v_a+v_b )\leq 0$ and $v_a-v_b>0$. Therefore, $(v_a-v_b )-(v_a^2-v_b^2 )\leq 0$, i.e., $(v_a-v_b )\leq(v_a^2-v_b^2)$.

There are three works \cite{R3S,R4S,R15Y} related to the proposed method, so next we would like to emphasize their differences. Several differences between our LAT-AD and  \cite{R3S,R4S} are listed as follows: we use clipped function and the local activity; the activity is calculated by the interval-updated way; and our method uses the division between gradient and local activity, but \cite{R3S,R4S} use the multiplication; our edge-stop function comes from Eq. (3), while the ones in \cite{R3S,R4S} use Eq. (4).
The differences between the proposed LAT-AD and \cite{R15Y} are listed as follows:
\begin{enumerate}
  \item The local activity is leveraged in our paper, which makes the relative impacts more efficient. The detailed operation of activity used in paper \cite{R15Y} is very complex and the window for their activity is often set to be larger than $3\times3$. In this paper, we aim to achieve fast depth filtering for distorted image compressed by HEVC coder \cite{R41J}, so we just use $3\times3$ window centered at pixel $D_i$ to get the 8-connected standard deviation $v_i$ instead of variance, because if variance is used, small variance can be easily dominated by large variance, and will have little contribution to the diffusion.
  \item A clipped function is used for local activity to make diffusion stable during anisotropic diffusion, because pixels with very large local activity render local activity-tuned anisotropic diffusion useless for pixels with smaller local activity measurement.
  \item During the iterative diffusion, the updated activity is used to control the degree of diffusion. If the image's diffusion is too fast, the fixed local activity often tends to blur the image discontinuities. The time-updated local activity can always preserve the sharp boundaries in the image, but it requires extra calculation of the local activity in every iteration. The interval-updated activity is a good alternative, especially when fast filtering is required by some applications.
\end{enumerate}
\section{Local activity-tuned relative total variation}
The classic total variation (TV) method \cite{R16L} can be written as:
\begin{equation}
E_{TV}(I|I^0)=\mathop{\arg\min}_{I} \iint_{\Omega} ||\nabla I||d_x d_y + \frac{\lambda}{2} \iint_{\Omega}(I-I^0)^2d_x d_y,
\end{equation}
where $\Omega$ is the domain of the image and $I^0$ is the initial image.

To compare with the anisotropic diffusion, according to \cite{R42E} the Euler-Langrage Equation of the TV model can be used, which is given as follows:
\begin{equation}
\lambda (I-I^0)-\nabla \cdot(\frac{\nabla I}{||\nabla I||})=0.
\end{equation}

Comparing Eq. (2) and Eq. (18), it is clear that the total variance model can be viewed as a special case of the anisotropic diffusion with edge-stop function to be  $\frac{1}{||\nabla I||}$.

In order to extract the main structure from the textured background, a relative total variation (RTV) model is proposed in \cite{R17L}, which is based on two variation measures. The first is the conventional windowed total variation (WTV) measure to capture visual saliency of the image:
\begin{equation}
\begin{split}
\mathcal{D}_x(p)=\sum_{q \in \mathbb{N}_p} g_{p,q} |(\partial_x I)_q| \\
\mathcal{D}_y(p)=\sum_{q \in \mathbb{N}_p} g_{p,q} |(\partial_y I)_q|,
\end{split}
\end{equation}
where $g_(p,q)$ is a Gaussian weighting function with variance $\sigma^2$,
\begin{equation}
g_{p,q}=exp(-\frac{(x_p-x_q^2)+(y_p-y_q)^2}{2\sigma^2}).
\end{equation}

In addition, a windowed inherent variation (WIV) measure is introduced in \cite{R17L} as follows:
\begin{equation}
\begin{split}
\mathcal{L}_x(p)=|\sum_{q \in \mathbb{N}_p} g_{p,q} (\partial_x I)_q| \\
\mathcal{L}_y(p)=|\sum_{q \in \mathbb{N}_p} g_{p,q} (\partial_y I)_q|.
\end{split}
\end{equation}

Note that it adds the variations rather than the absolute values of gradient. Therefore its response is much smaller in a window that only contains textures.

To further enhance the contrast between texture and structure, the ratio of the WTV and WIV, which is called the RTV regularizer, is used to remove textures from the image and only keep the structure \cite{R17L}. The overall objective function is
\begin{equation}
\mathop{\arg\min}_{I} \sum_{p}(I_p-I^0_p)^2+\sum_{p}\lambda (\frac{\mathcal{D}_x(p)}{\mathcal{L}_x(p)+\epsilon}+\frac{\mathcal{D}_y(p)}{\mathcal{L}_y(p)+\epsilon}),
\end{equation}
where $\epsilon$ is a small positive number to avoid dividing by zero.

Inspired by the RTV, we propose a local activity-tuned relative total variation for image smoothing (LAT-RTV), which is given by
\begin{equation}
\mathop{\arg\min}_{I} \sum_{p}(I_p-I^0_p)^2+\sum_{p}\lambda \frac{(\frac{\mathcal{D}_x(p)}{\mathcal{L}_x(p)+\epsilon}+\frac{\mathcal{D}_y(p)}{\mathcal{L}_y(p)+\epsilon})}{v_p}
\end{equation}
where the clipped and normalized local activity measurement $v_p$ is obtained according to Eq. (12-16).

Most pixels around edges have high activity. By dividing $v_p$ in Eq. (23), these pixels will have less contribution to the RTV term so that the edge will be preserved. Thus, compared to RTV \cite{R17L}, the proposed LAT-RTV in Eq. (23) will further smoothen the details and textures in the image, but will preserve the structural information.

Due to the non-convexity of Eq. (23), its solution cannot be directly obtained. As described in \cite{R17L,R34D}, an objective function with a quadratic term as penalty can be optimized linearly. According to \cite{R17L}, the LAT-RTV term can be decomposed into a quadratic part and a non-linear part. By putting Eq. (19) and Eq. (21) into the LAT-RTV term in the $x$-direction, it can be re-written as:

\begin{align}
\sum_{p}\frac{\frac{\mathcal{D}_x(p)}{\mathcal{L}_x(p)+\epsilon}}{v_p}
&=\sum_{p}\frac{\frac{\sum_{q \in \mathbb{N}_p} g_{p,q} \cdot |(\partial_x I)_p|}{\mathcal{L}_x(p)+\epsilon}}{v_p}\notag\\
&=\sum_{p}\sum_{q \in \mathbb{N}_p} \frac{\frac{g_{p,q} \cdot |(\partial_x I)_p|}{\mathcal{L}_x(p)+\epsilon}}{v_p}\notag\\
&\approx \sum_{p}\sum_{q \in \mathbb{N}_p} \frac{g_{p,q}}{\mathcal{L}_x(p)+\epsilon}\cdot \frac{1}{|(\partial_x I)_p|+\epsilon}\cdot \frac{1}{v_p}\cdot (\partial_x I)^2_p
\end{align}

This can be rewritten as
\begin{equation}
\sum_{p}\frac{\frac{\mathcal{D}_x(p)}{\mathcal{L}_x(p)+\epsilon}}{v_p}\approx \sum_{p} s_{x,p} \cdot c_p \cdot (\partial_x I)^2_p,
\end{equation}
where
\begin{equation}
c_p=\frac{1}{v_p}
\end{equation}
\begin{equation}
s_{x,p}=\sum_{q \in \mathbb{N}_p}\frac{g_{p,q}}{\mathcal{L}_x(p)+\epsilon} \cdot \frac{1}{|(\partial_x I)_p|+\epsilon}.
\end{equation}
Similarly, the LAT-RTV term in the $y$-direction can be written as:
\begin{equation}
\sum_{p}\frac{\frac{\mathcal{D}_y(p)}{\mathcal{L}_y(p)+\epsilon}}{v_p}\approx \sum_{p} s_{y,p} \cdot c_p \cdot (\partial_y I)^2_p
\end{equation}
where
\begin{equation}
s_{y,p}=\sum_{q \in \mathbb{N}_p}\frac{g_{p,q}}{\mathcal{L}_y(p)+\epsilon} \cdot \frac{1}{|(\partial_y I)_p|+\epsilon}
\end{equation}

For simplicity, we re-write Eq. (23) in the form of matrix as follows:
\begin{align}
&\mathop{\arg\min}_{\bm{I}} \bm{(V_I-V_{I_0})^T(V_I-V_{I_0})} + \notag\\
&\lambda \bm{((V_I)^T (G_x)^T S_x C G_x V_I+ V_I^T G_y^T S_y C G_y V_I)}
\end{align}

In Eq. (30), $\bm{V_I}$ and $\bm{V_{I_{0}}}$ are respectively the vector representation of $I$ and $I_0$, $\bm{G_x}$ and $\bm{G_y}$ are the Toeplitz matrices from the discrete gradient operators using forward difference. $\bm{S_x}$, $\bm{S_y}$, and $\bm{C}$ are the diagonal matrices, whose diagonal values are $\bm{S_x}[i,i]=s_{x,i}$, $\bm{S_y}[i,i]=s_{y,i},$ and $\bm{C}[i,i]=c_i$.

To minimize Eq. (30), we take the derivative with respect to $\bm{V_I}$ and the solution can be written as:
\begin{equation}
\bm{V_{I_0}}=(\bm{E}+\lambda(\bm{{G_x}^T {S_x} C G_x}+\bm{G_y^T {S_y} C G_y}) \cdot \bm{V_I} )
\end{equation}
where $\bm{E}$ is the identity matrix.

Finally, given the initial image $I_0$, the detailed iterative optimization procedure of LAT-RTV is presented as follows:
\begin{enumerate}
  \item In each iteration, use Eq. (27) and Eq. (29) to calculate $s_x$ and $s_y$ in order to get matrices $\bm{S_x}$ and $\bm{S_y}$. In the first iteration, $\bm{S_x^0}$ and  $\bm{S_y^0}$ are obtained from $I_0$, otherwise $\bm{S_x}^t$ and $\bm{S_y^t}$ are obtained from $I_t$, which in the form of vector is $\bm{V_{I_t}}$.
  \item Given $\bm{S_x^{t-1}}$, $\bm{S_y^{t-1}}$, $\bm{G_x}$, and $\bm{G_y}$,  the vector results can be obtained in each iteration as follows, according to Eq. (32).
  \item After $\aleph$ times iterations with step (1-2), $\bm{V_{I_t}}$ is re-arranged into a matrix $\bm{I_t}$ with size $M\times N$, which is the final output image.
\end{enumerate}
\begin{equation}
\bm{V_{I_t}}=(\bm{E}+\lambda(\bm{{G_x}^T {S_x^{t-1}} C G_x}+\bm{G_y^T {S_y^{t-1}} C G_y}))^{-1} \cdot\ \bm{V_{I_{t-1}}} 
\end{equation}

In contrast to LAT-RTV, the product between $v_p$ and the RTV is firstly proposed to achieve image denoising (denoted as LAT-RTVd) as follows:
\begin{equation}
\mathop{\arg\min}_{I} \sum_{p}(I_p-I^0_p)^2+\sum_{p}\lambda (\frac{\mathcal{D}_x(p)}{\mathcal{L}_x(p)+\epsilon}+\frac{\mathcal{D}_y(p)}{\mathcal{L}_y(p)+\epsilon})\cdot v_p
\end{equation}

The solution for LAT-RTVd of Eq. (33) can be obtained similarly to the derivation for LAT-RTV, which is presented in Eq. (34). Here, $\bm{W}$ is the diagonal matrix and its $p$-th diagonal value is $v_p$.
\begin{equation}
\bm{V_{I_t}}=(\bm{E}+\lambda(\bm{{G_x}^T {S_x^{t-1}} W G_x}+\bm{G_y^T {S_y^{t-1}} W G_y}))^{-1} \cdot \bm{V_{I_{t-1}}} 
\end{equation}

Just as the denoising of LAT-AD, because the product of RTV and normalized and clipped standard variation can capture the locations of the noises in the contaminated image, LAT-RTVd can smoothen the detected noisy pixels to achieve image denoising. This comes from the fact that gradient information has noise's gradient change except for boundary information change, but local variance or standard deviation is usually a stable statistic feature for image without obvious noises.

In the RTV model, whether a pixel is judged as a texture pixel or a structural pixel depends on the gradient changes of local information within a patch through the WTV and WIV. Thus, the RTV model smoothens all the textural pixels so as to extract structure from texture. However, our LAT-RTVd judges whether and how much a pixel belongs to a noisy pixel based on local activity-tuned RTV, so LAT-RTVd prefers to smoothen noisy pixels detected by local activity and gradient, rather than all the textural pixels. Therefore, our LAT-RTVd has the ability to maintain more detailed textural information than RTV.

From Eq. (25-27), it can be clearly seen that LAT-RTV employs the multiplication between local activity $v_p$ and gradient $|(\partial_x I)_p|$ in the $x$-direction, but it is in the way of division between RTV and normalized clipped local activity. On the contrary, LAT-RTVd uses the division of $v_p$ and $|(\partial_x I)_p|$ in the $x$-direction.
\begin{figure}[!t]
\centering
\includegraphics[width=3in]{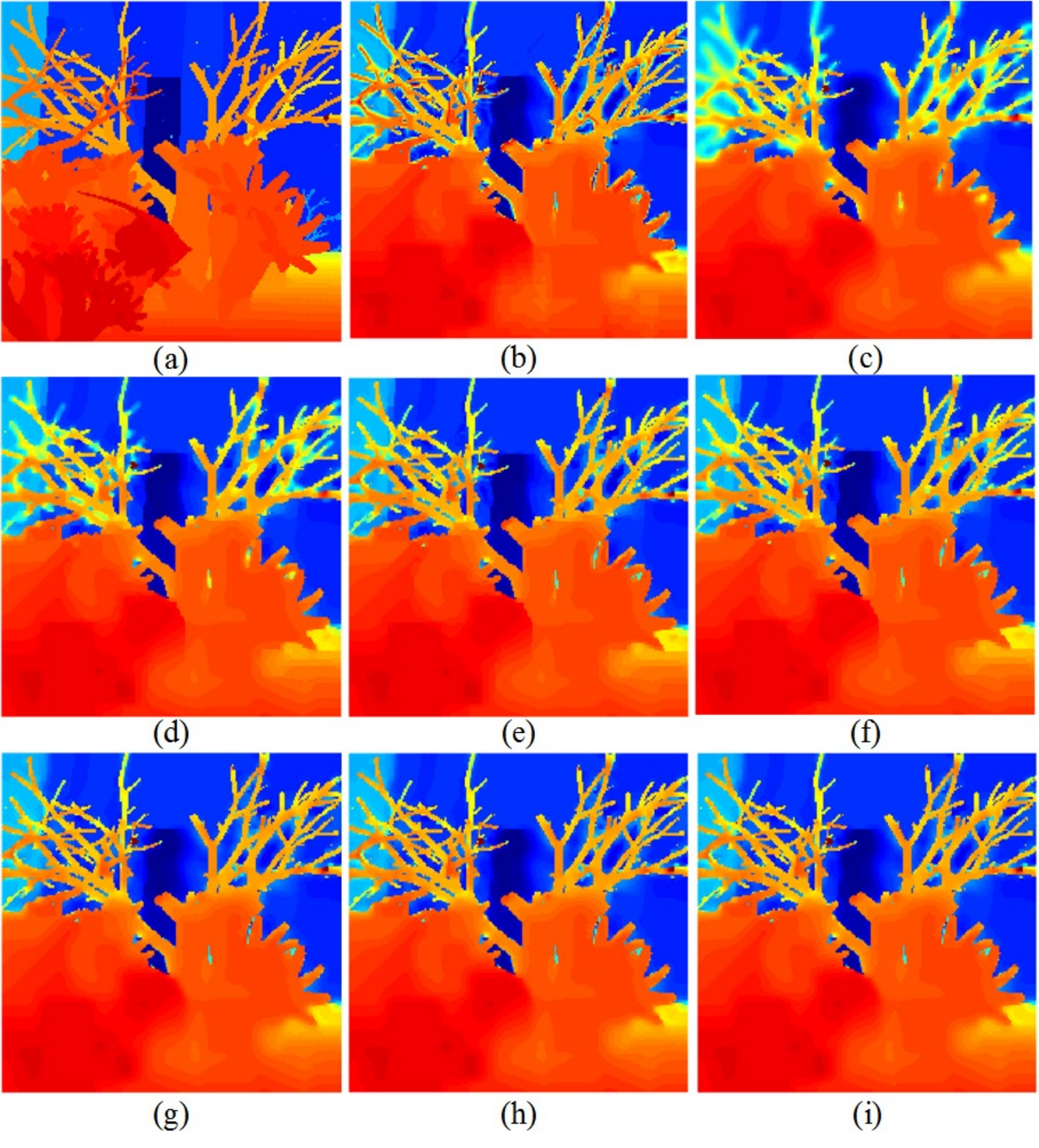}
\caption{(a) Part of the first frame depth image of Shark with QP=39, (b) compressed by HEVC, (c) PM diffusion for (b), (d) FLAT-AD, (e) PLAT-AD, (f) TLAT-AD, (h) FLAT-AD (I) , (h) PLAT-AD (I) , (h) TLAT-AD.}
\label{Fig1}
\end{figure}
\begin{figure}[!t]
\centering
\includegraphics[width=2.5in]{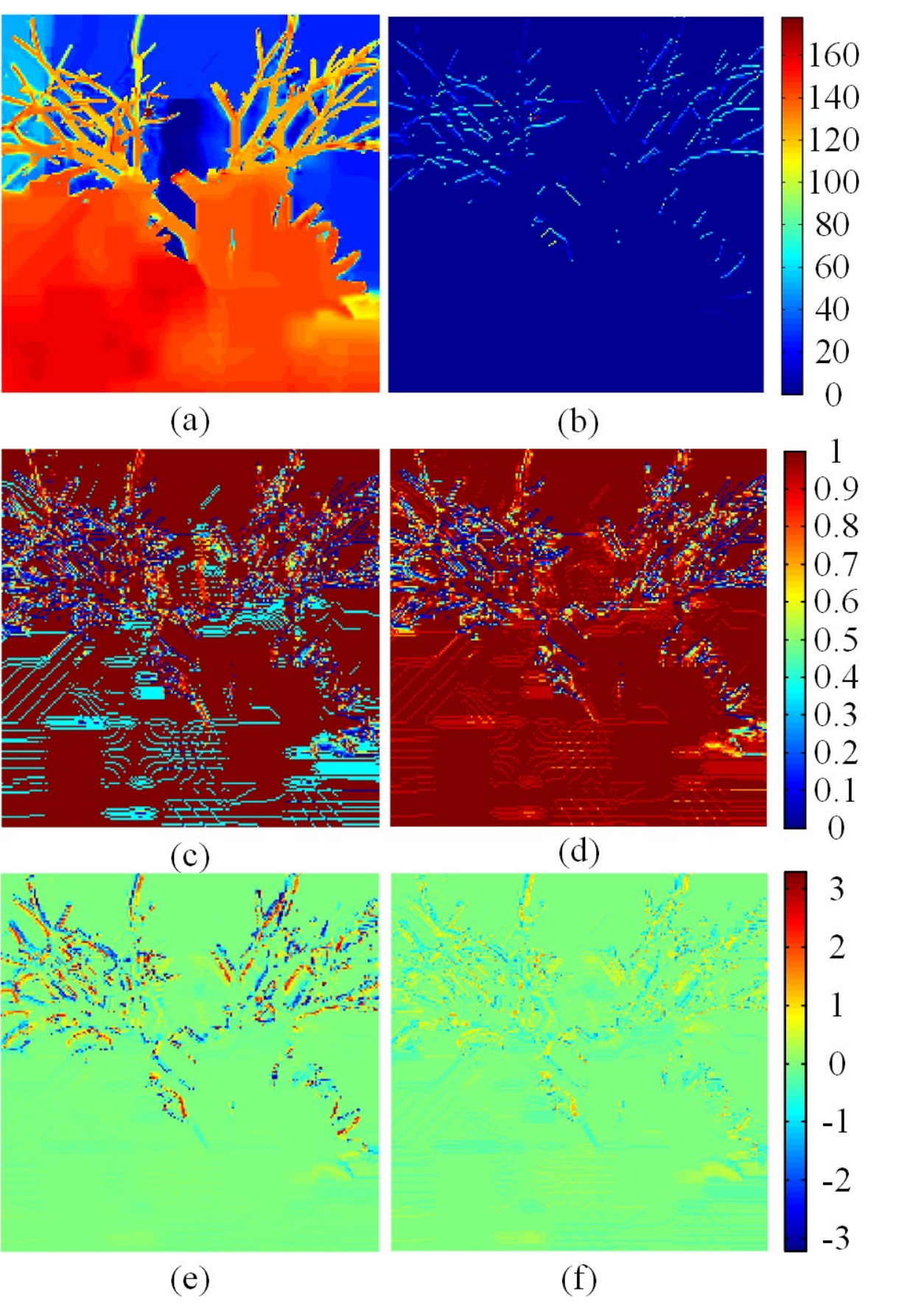}
\caption{(a) Part of the first frame depth image of Shark with QP=39, (b) the gradient in the vertical direction, (c) the edge-stop function output  with Eq. (10),  (d) the edge-stop function output  with Eq. (11), (e) the first step diffusion in the vertical direction using Eq. (10), (f) the first step diffusion in the vertical direction using Eq. (11).}
\label{Fig2}
\end{figure}
\section{Experimetal results and analysis}
\begin{table}[!t]
\renewcommand{\arraystretch}{1.3}
\caption{The objective quality comparison for depth images filtered by different methods when QP=37, 39, 41}
\label{table1}
\tiny
\centering
\begin{tabular}{|c|c|c|c|c|c|c|c|c|c|}
\hline
       M/Seq &        U-1 &        S-1 &       C-37 &       B-10 &        U-5 &        S-5 &       C-39 &        B-8 & {\bf Ave.} \\
\hline
   Coded41 &     44.23  &     39.56  &     40.60  &     37.85  &     44.19  &     39.46  &     40.58  &     37.79  &     40.53  \\
\hline
  CVBF\cite{R14L} &     44.27  &     39.33  &     40.45  &     37.41  &     44.27  &     39.25  &     40.40  &     37.27  &     40.33  \\
\hline
  ADTF\cite{R13X} &     44.18  &     39.38  &     40.45  &     37.35  &     44.16  &     39.30  &     40.43  &     37.31  &     40.32  \\
\hline
   TSF\cite{R33L} &     44.34  &     38.87  &     40.49  &     37.50  &     44.32  &     38.78  &     40.41  &     37.36  &     40.26  \\
\hline
    FLAT-AD &     44.64  &     39.50  &     40.65  &     37.71  &     44.63  &     39.42  &     40.61  &     37.64  &     40.60  \\
\hline
    TLAT-AD &     44.58  &     39.60  &     40.71  &     37.65  &     44.56  &     39.52  &     40.70  &     37.59  &     40.61  \\
\hline
    PLAT-AD &     44.64  &     39.61  &     40.71  &     37.72  &     44.62  &     39.53  &     40.68  &     37.66  &     40.65  \\
\hline
FLAT-AD (I) &     44.82  &     39.75  &     40.87  &     38.07  &     44.80  &     39.66  &     40.93  &     37.98  & {\bf 40.86 } \\
\hline
PLAT-AD (I) &     44.81  &     39.74  &     40.86  &     38.05  &     44.78  &     39.66  &     40.93  &     37.97  & {\bf 40.85 } \\
\hline
                                                                                                        \multicolumn{ 10}{|c|}{} \\
\hline
   Coded39 &     45.73  &     40.85  &     42.12  &     38.93  &     45.71  &     40.77  &     42.14  &     38.83  &     41.89  \\
\hline
  CVBF\cite{R14L} &     45.88  &     40.86  &     42.09  &     38.66  &     45.86  &     40.78  &     42.16  &     38.52  &     41.85  \\
\hline
  ADTF\cite{R13X} &     45.68  &     40.68  &     41.98  &     38.46  &     45.64  &     40.60  &     42.02  &     38.37  &     41.68  \\
\hline
   TSF\cite{R33L} &     45.87  &     39.91  &     41.97  &     38.50  &     45.84  &     39.81  &     41.95  &     38.33  &     41.52  \\
\hline
    FLAT-AD &     46.21  &     40.71  &     42.22  &     37.79  &     46.20  &     40.64  &     42.31  &     38.71  &     41.85  \\
\hline
    TLAT-AD &     46.16  &     40.96  &     42.18  &     37.65  &     46.14  &     40.89  &     42.25  &     37.59  &     41.73  \\
\hline
    PLAT-AD &     46.21  &     40.93  &     42.24  &     38.82  &     46.20  &     40.86  &     42.32  &     38.74  &     42.04  \\
\hline
FLAT-AD (I) &     46.42  &     41.10  &     42.46  &     39.16  &     46.42  &     41.03  &     42.59  &     39.06  & {\bf 42.28 } \\
\hline
PLAT-AD (I) &     46.41  &     41.10  &     42.45  &     39.15  &     46.41  &     41.02  &     42.58  &     39.05  & {\bf 42.27 } \\
\hline
                                                                                                        \multicolumn{ 10}{|c|}{} \\
\hline
   Coded37 &     47.30  &     42.22  &     43.77  &     40.09  &     47.28  &     42.15  &     43.89  &     40.10  &     43.35  \\
\hline
  CVBF\cite{R14L} &     47.46  &     42.14  &     43.72  &     39.68  &     47.44  &     42.08  &     43.78  &     39.79  &     43.26  \\
\hline
  ADTF\cite{R13X} &     47.23  &     41.95  &     43.57  &     39.60  &     47.20  &     41.89  &     43.65  &     39.63  &     43.09  \\
\hline
   TSF\cite{R33L} &     47.47  &     40.86  &     43.52  &     39.40  &     47.46  &     40.74  &     43.54  &     39.54  &     42.82  \\
\hline
    FLAT-AD &     47.89  &     42.22  &     43.94  &     39.95  &     47.88  &     42.15  &     44.08  &     39.96  &     43.51  \\
\hline
    TLAT-AD &     47.85  &     42.35  &     43.93  &     39.96  &     47.84  &     42.29  &     44.05  &     39.96  &     43.53  \\
\hline
    PLAT-AD &     47.91  &     42.31  &     43.97  &     40.01  &     47.90  &     42.24  &     44.10  &     40.01  &     43.56  \\
\hline
FLAT-AD (I) &     48.12  &     42.48  &     44.10  &     40.35  &     48.10  &     42.42  &     44.31  &     40.31  & {\bf 43.77 } \\
\hline
PLAT-AD (I) &     48.11  &     42.48  &     44.10  &     40.34  &     48.09  &     42.42  &     44.31  &     40.30  & {\bf 43.77 } \\
\hline
\end{tabular}
\end{table}

\begin{table}[!t]
\renewcommand{\arraystretch}{1.3}
\caption{The objective quality comparison for depth images filtered by different methods when QP=31, 33, 35}
\label{table2}
\tiny
\centering
\begin{tabular}{|c|c|c|c|c|c|c|c|c|c|}
\hline
       M/Seq &        U-1 &        S-1 &       C-37 &       B-10 &        U-5 &        S-5 &       C-39 &        B-8 & {\bf Ave.} \\
\hline
   Coded35 &     48.91  &     43.62  &     45.29  &     41.43  &     48.87  &     43.54  &     45.46  &     41.34  &     44.81  \\
\hline
  CVBF\cite{R14L} &     49.08  &     43.39  &     45.16  &     41.03  &     49.05  &     43.30  &     45.12  &     40.81  &     44.62  \\
\hline
  ADTF\cite{R13X} &     48.79  &     43.17  &     44.99  &     40.91  &     48.76  &     43.09  &     44.98  &     40.79  &     44.44  \\
\hline
   TSF\cite{R33L} &     49.16  &     41.89  &     45.22  &     41.05  &     49.14  &     41.76  &     45.17  &     40.73  &     44.27  \\
\hline
    FLAT-AD &     49.68  &     43.83  &     45.71  &     41.62  &     49.65  &     43.76  &     45.79  &     41.53  &     45.20  \\
\hline
    TLAT-AD &     49.71  &     43.89  &     45.67  &     41.60  &     49.68  &     43.82  &     45.75  &     41.51  &     45.20  \\
\hline
    PLAT-AD &     49.69  &     43.89  &     45.69  &     41.61  &     49.66  &     43.81  &     45.77  &     41.52  &     45.21  \\
\hline
FLAT-AD (I) &     49.87  &     43.97  &     45.81  &     41.85  &     49.85  &     43.90  &     45.95  &     41.74  & {\bf 45.37 } \\
\hline
PLAT-AD (I) &     49.86  &     43.98  &     45.81  &     41.85  &     49.84  &     43.90  &     45.94  &     41.73  & {\bf 45.36 } \\
\hline
                                                                                                        \multicolumn{ 10}{|c|}{} \\
\hline
   Coded33 &     50.33  &     44.97  &     46.73  &     42.60  &     50.29  &     44.90  &     46.81  &     42.72  &     46.17  \\
\hline
  CVBF\cite{R14L} &     50.50  &     44.52  &     46.45  &     41.88  &     50.50  &     44.42  &     46.20  &     42.16  &     45.83  \\
\hline
  ADTF\cite{R13X} &     50.19  &     44.14  &     46.22  &     41.89  &     50.13  &     44.22  &     46.00  &     42.06  &     45.61  \\
\hline
   TSF\cite{R33L} &     50.62  &     42.49  &     46.56  &     41.78  &     50.70  &     42.35  &     46.22  &     42.19  &     45.36  \\
\hline
    FLAT-AD &     51.28  &     45.09  &     47.09  &     42.74  &     51.24  &     45.16  &     46.98  &     42.87  &     46.56  \\
\hline
    TLAT-AD &     51.32  &     45.24  &     47.05  &     42.72  &     51.27  &     45.17  &     46.92  &     42.85  &     46.57  \\
\hline
    PLAT-AD &     51.30  &     45.16  &     47.07  &     42.74  &     51.26  &     45.23  &     46.95  &     42.86  &     46.57  \\
\hline
FLAT-AD (I) &     51.51  &     45.33  &     47.25  &     43.11  &     51.46  &     45.26  &     47.30  &     42.97  & {\bf 46.77 } \\
\hline
PLAT-AD (I) &     51.50  &     45.33  &     47.25  &     43.11  &     51.46  &     45.26  &     47.30  &     42.97  & {\bf 46.77 } \\
\hline
                                                                                                        \multicolumn{ 10}{|c|}{} \\
\hline
   Coded31 &     51.71  &     46.35  &     48.28  &     43.95  &     51.65  &     46.28  &     48.35  &     43.84  &     47.55  \\
\hline
  CVBF\cite{R14L} &     51.92  &     45.56  &     47.76  &     43.16  &     51.88  &     45.47  &     47.34  &     42.90  &     47.00  \\
\hline
  ADTF\cite{R13X} &     51.38  &     45.17  &     47.43  &     43.00  &     51.45  &     45.09  &     47.10  &     42.82  &     46.68  \\
\hline
   TSF\cite{R33L} &     51.95  &     42.93  &     47.87  &     43.16  &     52.01  &     42.76  &     47.37  &     42.78  &     46.35  \\
\hline
    FLAT-AD &     52.82  &     46.53  &     48.48  &     44.01  &     52.77  &     46.47  &     48.30  &     43.86  &     47.91  \\
\hline
    TLAT-AD &     52.83  &     46.63  &     48.45  &     43.99  &     52.78  &     46.57  &     48.25  &     43.85  &     47.92  \\
\hline
    PLAT-AD &     52.84  &     46.62  &     48.47  &     44.01  &     52.79  &     46.56  &     48.28  &     43.86  &     47.93  \\
\hline
FLAT-AD (I) &     53.01  &     46.70  &     48.62  &     44.22  &     52.96  &     46.64  &     48.62  &     44.07  & {\bf 48.11 } \\
\hline
PLAT-AD (I) &     53.02  &     46.71  &     48.63  &     44.22  &     52.96  &     46.65  &     48.63  &     44.07  & {\bf 48.11 } \\
\hline
\end{tabular}
\end{table}
In this section, we present extensive results to demonstrate the performance of the proposed methods. First, we first apply the proposed LAT-AD to the problem of artifact removal of piece-wise smooth image, such as depth image and clip-art image. Secondly, we validate the efficiency of the proposed LAT-RTV on image smoothing. Finally, our LAT-RTVd is compared with several denoising methods to demonstrate the novelty of the proposed method.
\subsection{Compressed depth image filtering with LAT-AD}
\begin{figure}[!t]
\centering
\includegraphics[width=3.5in]{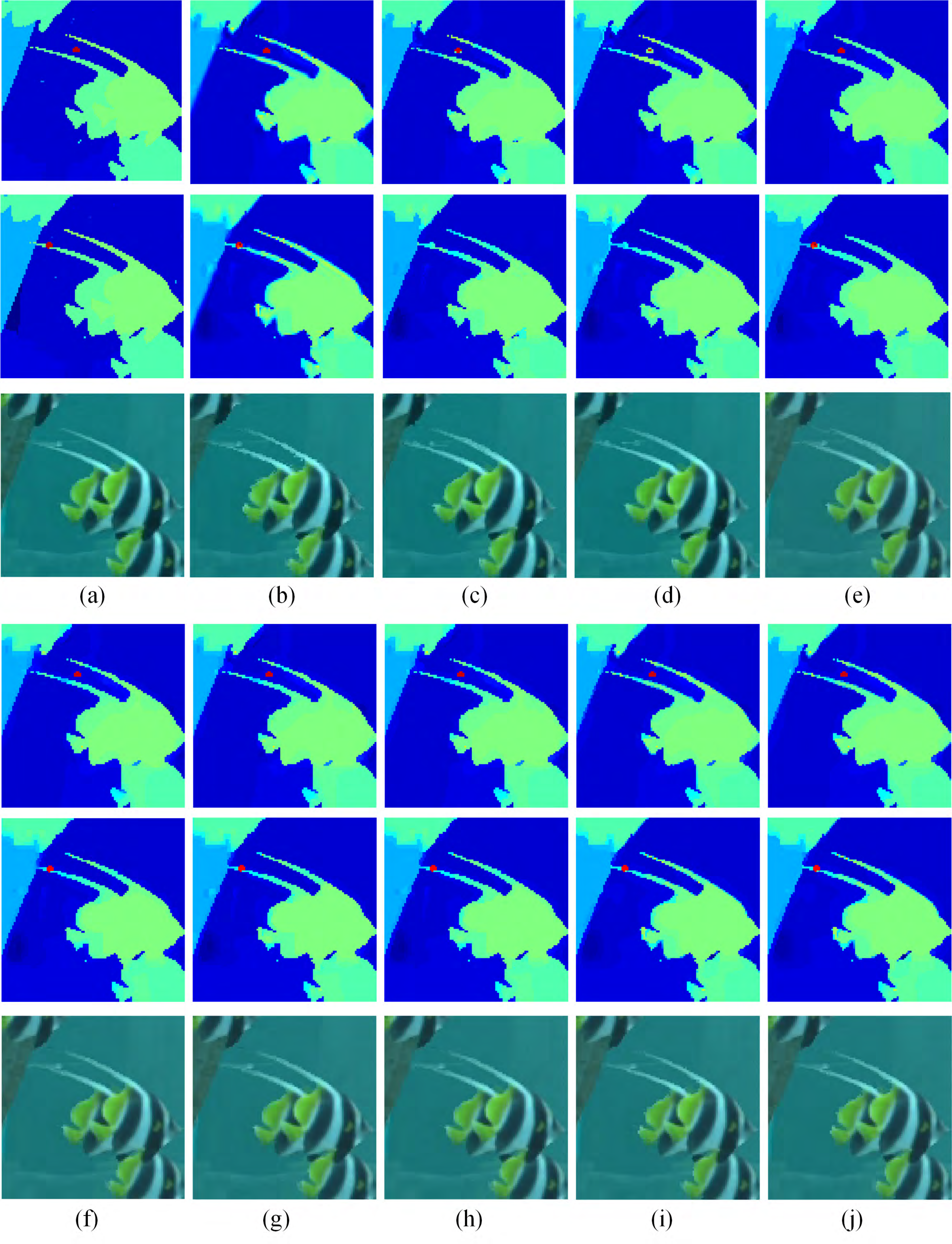}
\caption{The first row: (a) is part of the original depth map Shark in view 1, (b) HEVC (QP41), (c) CVBF, (d) ADTF, (e) TSF , (f-j) FLAT-AD, TLAT-AD, PLAT-AD, FLAT-AD (I), PLAT-AD (I); the second row of (a-j) is corresponding depth image of view 5; the third row of (a-j) is middle virtual images synthesized by corresponding depth image in the first and second row.}
\label{Fig3}
\end{figure}

\begin{figure*}[!t]
\centering
\includegraphics[width=5in]{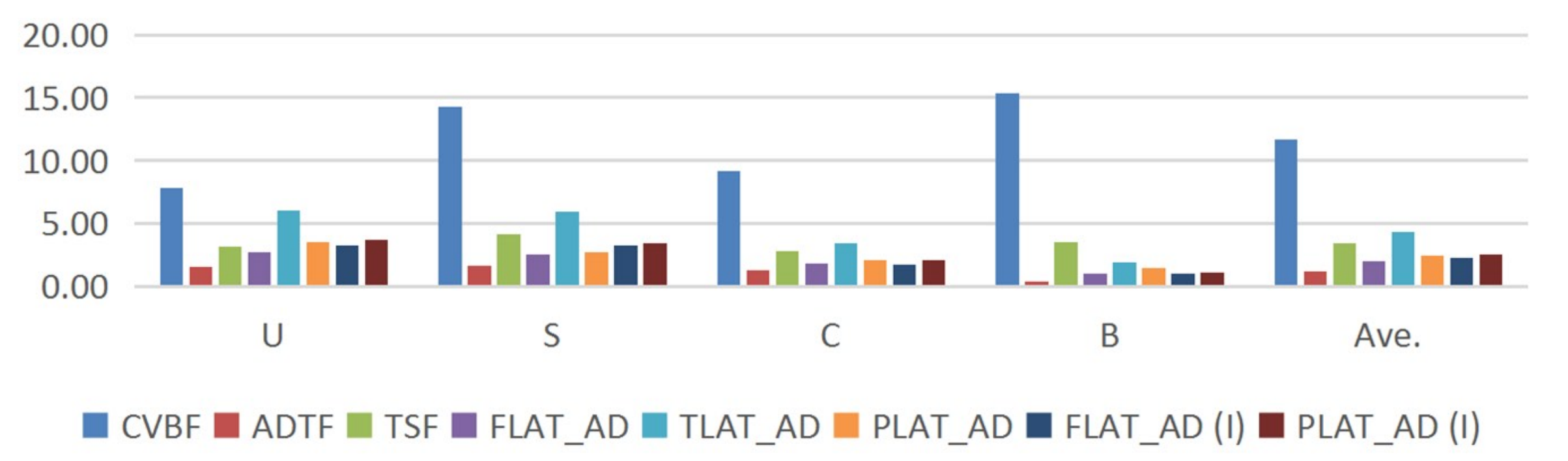}
\caption{The comparison of filtering time (seconds/frame) with different methods for compressed depth image.}
\label{Fig4}
\end{figure*}
\begin{figure*}[!t]
\centering
\includegraphics[width=7in]{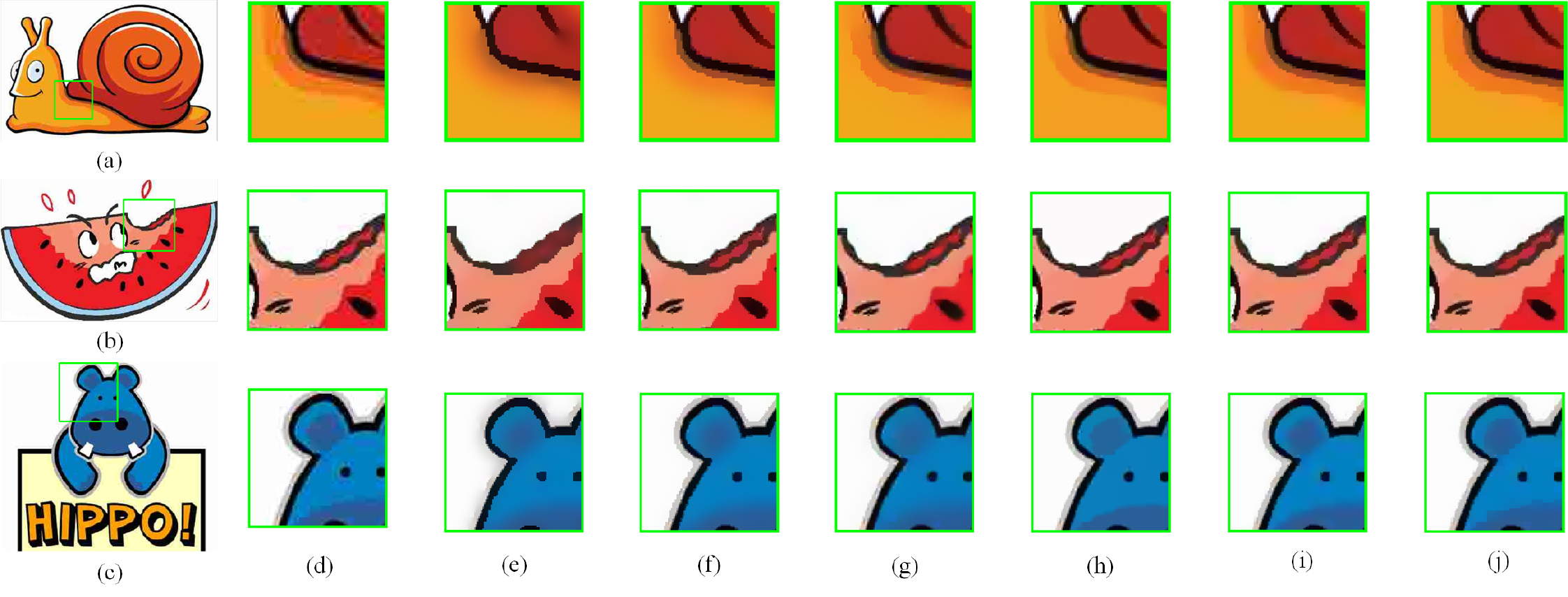}
\caption{The comparison of clip-art compression artifact removal with different methods: (a-c) Three noisy clip-art image; (d) is the boxed region from (a); (e) is filtered with TV [16] when iteration to be 50, K to be 20, $\lambda$ to be 0.25; (f) is filtered with TV [16] when iteration to be 50, $\rho$ to be 10, $\lambda$ to be 0.25; (g) is filtered by Modified TV [3] when iteration to be 100, ρ to be 25, $\lambda$ to be 0.25; (h) is filtered with L0 gradient minimization method [18] when lamda=0.01; (i) is filtered with FLAT-AD (I) when iteration to be 11, $(\rho_2)^2$ to be 17, $\lambda$ to be 0.25; (j) is filtered with FLAT-AD when iteration to be 50, $\rho_1$ to be 20, $\lambda$ to be 0.25.}
\label{Fig5}
\end{figure*}

\begin{figure}[!t]
\centering
\includegraphics[width=3.5in]{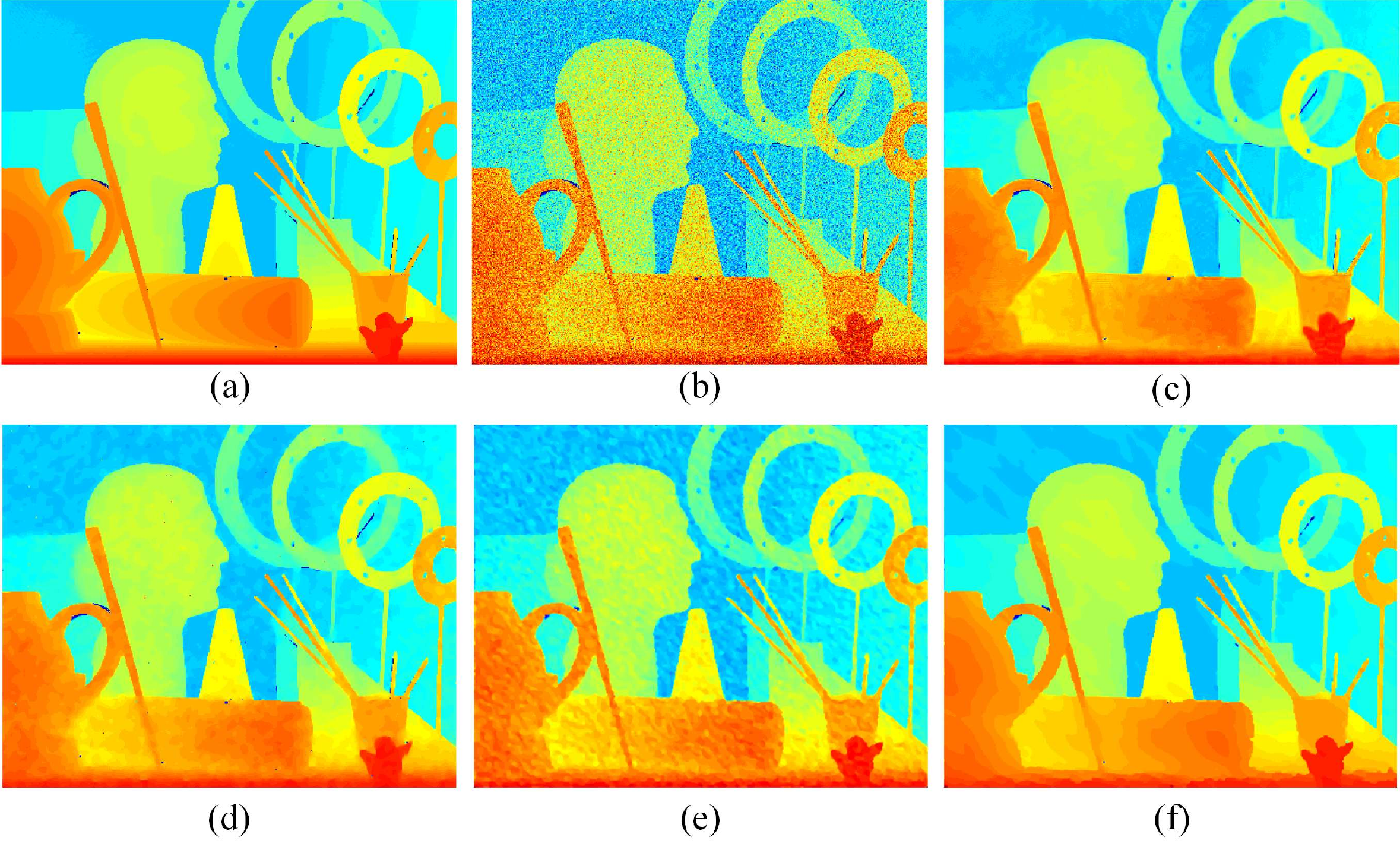}
\caption{The visual comparison of noisy disparity filtered by several methods, (a) Disparity image of Art; (b) Art with noise standard deviation to be 20; (c) BM3D [20]\cite{R20K}; (d) NLGBT \cite{R19W}; (e) RTV \cite{R17L}; (f)LAT-RTV.}
\label{Fig6}
\end{figure}
\begin{figure*}[!t]
\centering
\includegraphics[width=7in]{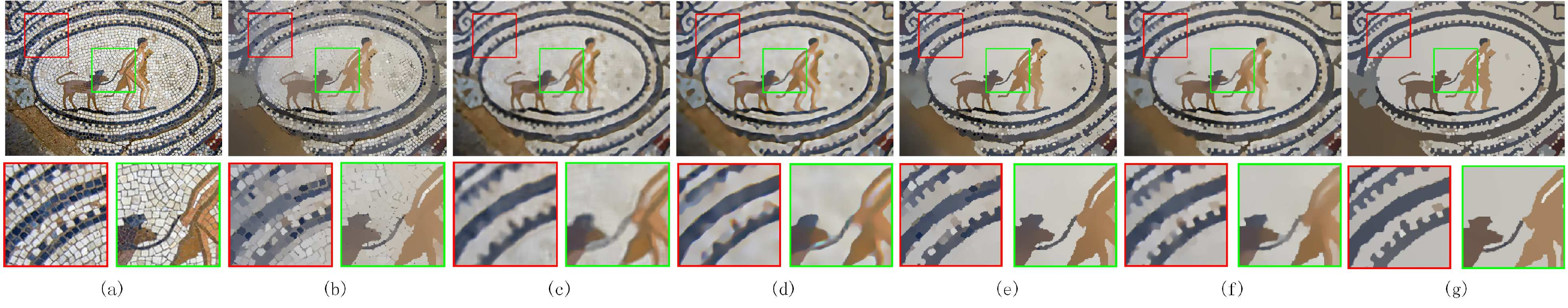}
\caption{(a) input image, (b-g) the visual comparison of five image smoothing methods including: WLS \cite{R21L}, RC \cite{R24L}, RGF \cite{R25Q}, RGIF \cite{R26B}, RTV \cite{R17L}, and our LAT-RTV.}
\label{Fig7}
\end{figure*}
\begin{figure}[!t]
\centering
\includegraphics[width=3.5in]{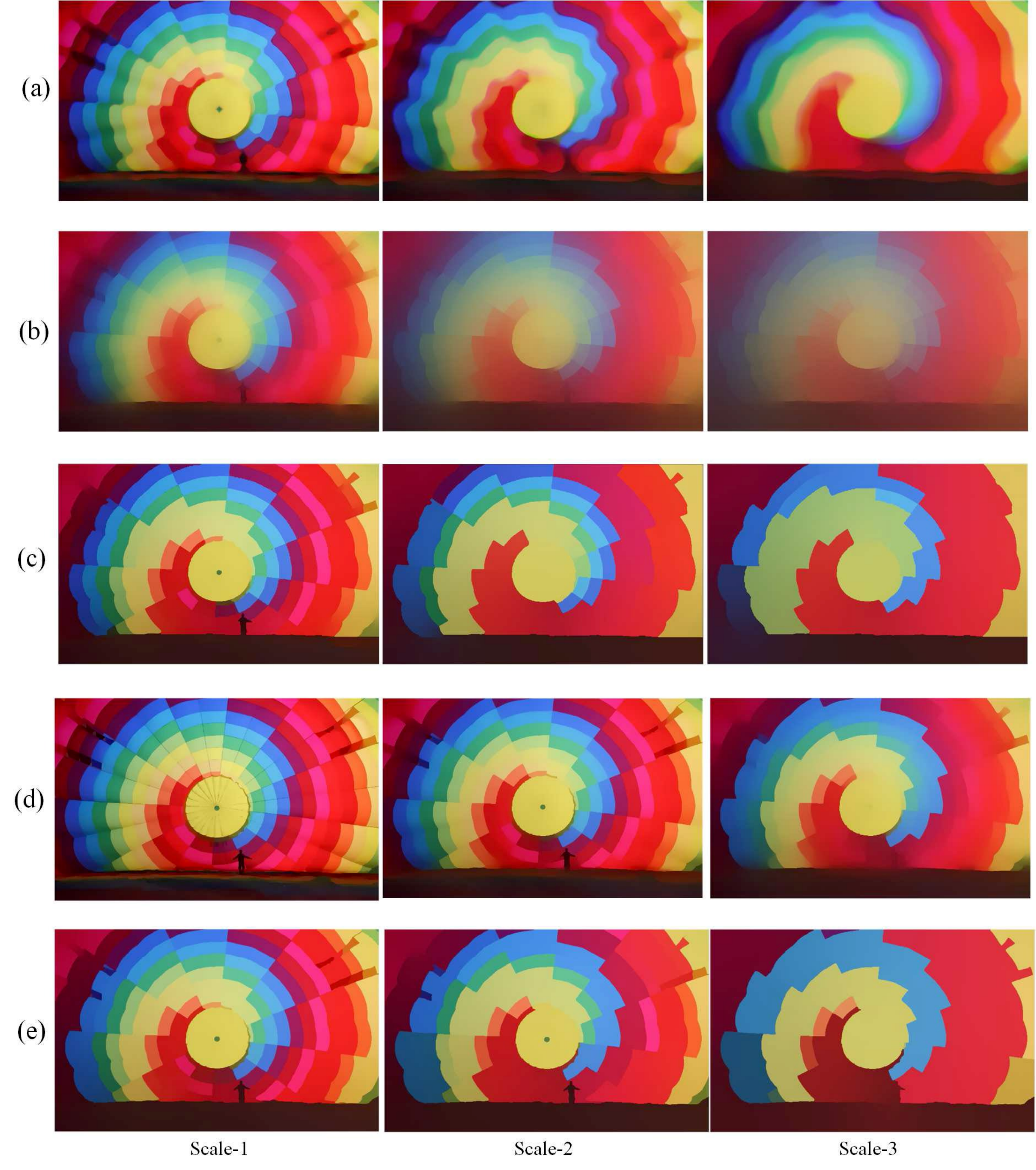}
\caption{Examples of the scale space dealt with five methods including: (a) RGF \cite{R25Q}; (b) WLS \cite{R21L}, (c) RGIF \cite{R26B}, (d) RTV \cite{R27K}, (e) LAT-RTV}
\label{Fig8}
\end{figure}

The depth maps are compressed by HEVC v16.8 \cite{R37D} with quantization parameter chosen as 31, 33, 35, 37, 39 and 41, respectively. We use four standard multi-view-plus-depth sequences: Nokia's Undo\_Dancer (U), NICT's Shark (S), Nagoya University's Champagne\_Tower (C) (in which the first 250 frames of these three sequences are tested) and HHI's Book\_Arrival (B) (the whole sequences with 100 frames are tested) \cite{R38I}. In the simulations, the 1D-fast mode of 3D-HEVC (HTM-DEV-2.0-dev3 version) \cite{R39J} is used to synthesize the virtual middle view using two views of uncompressed texture images and compressed depth images (filtered or non-filtered). In our experiment, all the sequences are set with the same parameters for filtering. For FLAT-AD, TLAT-AD, PLAT-AD, FLAT-AD (I), and PLAT-AD (I), $\lambda$ is 0.25, $h$ is 30 and the number of iteration is 11 when QP lower than 37, otherwise the number of iteration is 21, which are the experimental values. For FLAT-AD, TLAT-AD, and PLAT-AD, $\rho_1$  is set to be 30, while $\rho_2^2$ is 300 for FLAT-AD (I), and PLAT-AD (I).  For PLAT-AD and PLAT-AD (I), the interval is 5 when QP =31, 34, 35, but the interval $l$  is set to be 10 if QP=37, 39, 41.

The filtering results of the proposed method are compared with those of ADTF \cite{R13X}, CVBF \cite{R14L}, and TSF \cite{R33L}. For both filtered depth images and the synthesized virtual view (the middle view of two reference views), the peak signal noise ratio (PSNR) is taken as the objective evaluation of filtered depth images and corresponding synthesized images. The average PSNRs of different sequences are presented in TABLE-I, TABLE-II, and TABLE-III, where U-1 represents the view-1 of Undo\_Dancer (U), the notations of other sequences are defined similarly, and M/Seq denotes Method/Sequence.

From Fig.1 (d-f), it can be observed that the performances of FLAT-AD and TLAT-AD as well as PLAT-AD are different and the sharpness of TLAT-AD is stronger than PLAT-AD, but TLAT-AD requires updated activity information every time, so TLAT-AD has more complexity than PLAT-AD. The diffusion of FLAT-AD leads to blur of depth image's discontinuities, so it has the worst performance on boundary regions as compared with the other methods.  Different from FLAT-AD, TLAT-AD, and PLAT-AD, the performances of FLAT-AD (I), TLAT-AD (I), and PLAT-AD (I) are very similar, as shown in Fig. 1 (g-i). The reason is that the form of $(\frac{||\nabla I^t_{ij}||}{\rho_1 K^t_i})^2$ leads to more diffusion for some artifact pixels than the form of $\frac{||\nabla I^t_{ij}||^2}{\rho_2^2 K^t_i}$ during each iteration, as shown in Fig. 2 (e-f). The stop-function in Eq. (10) is more efficient to smoothen image, as compared to the stop-function with Eq. (11). But the stop-function of Eq. (11) in the proposed FLAT-AD (I), TLAT-AD (I), and PLAT-AD (I) does not change depth structures too much and most of detailed geometry information is well preserved during removing severe coding artifacts.
\begin{table*}[!t]
\renewcommand{\arraystretch}{1.3}
\caption{The objective quality comparison of synthesizing the virtual view for different sequences}
\label{table3}
\scriptsize
\centering
\begin{tabular}{|c|c|c|c|c|c|c|c|c|c|c|c|}
\hline
       M/Seq &          U &          S &          C &          B &       Ave. &        M/Seq &          U &          S &          C &          B & {\bf Ave.} \\
\hline
{\bf Coded41} &     49.30  &     48.20  &     46.60  &     51.34  &     48.86  & {\bf Coded39} &     49.89  &     49.08  &     47.77  &     52.14  &     49.72  \\
\hline
  CVBF\cite{R14L} &     50.99  &     50.02  &     47.44  &     52.66  &     50.28  &   CVBF\cite{R14L} &     51.56  &     50.90  &     48.31  &     53.35  & {\bf 51.03 } \\
\hline
  ADTF\cite{R13X} &     50.82  &     50.16  &     47.29  &     52.29  &     50.14  &   ADTF\cite{R13X} &     51.61  &     51.03  &     48.29  &     53.06  &     51.00  \\
\hline
   TSF\cite{R33L} &     50.86  &     49.87  &     47.47  &     52.71  &     50.23  &        TSF\cite{R33L} &     51.71  &     50.81  &     48.39  &     53.38  & {\bf 51.07 } \\
\hline
    FLAT-AD &     50.37  &     49.54  &     47.38  &     52.64  &     49.98  &     FLAT-AD &     51.20  &     50.46  &     48.37  &     53.39  &     50.86  \\
\hline
    TLAT-AD &     50.22  &     49.55  &     47.21  &     52.52  &     49.88  &     TLAT-AD &     51.06  &     50.52  &     48.27  &     53.30  &     50.79  \\
\hline
    PLAT-AD &     50.32  &     49.55  &     47.31  &     52.55  &     49.93  &     PLAT-AD &     51.11  &     50.55  &     48.32  &     53.30  &     50.82  \\
\hline
FLAT-AD (I) &     50.42  &     49.58  &     49.17  &     52.41  & {\bf 50.40 } & FLAT-AD (I) &     51.29  &     50.64  &     48.43  &     53.24  &     50.90  \\
\hline
PLAT-AD (I) &     50.39  &     49.57  &     49.17  &     52.43  & {\bf 50.39 } & PLAT-AD (I) &     51.25  &     50.64  &     48.43  &     53.25  &     50.89  \\
\hline
                                                                                                                                  \multicolumn{ 12}{|c|}{} \\
\hline
{\bf Coded37} &     50.71  &     50.00  &     48.53  &     53.04  &     50.57  & {\bf Coded35} &     51.47  &     50.93  &     49.37  &     53.91  &     51.42  \\
\hline
  CVBF\cite{R14L} &     52.57  &     51.67  &     48.96  &     54.08  & {\bf 51.82 } &   CVBF\cite{R14L} &     53.47  &     52.45  &     49.66  &     54.76  & {\bf 52.59 } \\
\hline
  ADTF\cite{R13X} &     52.46  &     51.78  &     48.97  &     53.92  &     51.78  &   ADTF\cite{R13X} &     53.27  &     52.45  &     49.66  &     54.68  & {\bf 52.52 } \\
\hline
   TSF\cite{R33L} &     52.73  &     51.66  &     48.96  &     54.06  & {\bf 51.85 } &        TSF\cite{R33L} &     53.20  &     52.29  &     49.75  &     54.79  &     52.51  \\
\hline
    FLAT-AD &     52.18  &     51.27  &     49.03  &     54.21  &     51.67  &     FLAT-AD &     52.80  &     52.09  &     49.75  &     54.83  &     52.37  \\
\hline
    TLAT-AD &     52.03  &     51.41  &     48.96  &     54.12  &     51.63  &     TLAT-AD &     52.71  &     52.11  &     49.71  &     54.78  &     52.33  \\
\hline
    PLAT-AD &     52.09  &     51.39  &     49.00  &     54.16  &     51.66  &     PLAT-AD &     52.73  &     52.12  &     49.72  &     54.81  &     52.35  \\
\hline
FLAT-AD (I) &     52.23  &     51.66  &     49.17  &     54.17  &     51.81  & FLAT-AD (I) &     52.81  &     52.23  &     49.84  &     54.86  &     52.44  \\
\hline
PLAT-AD (I) &     52.22  &     51.65  &     49.17  &     54.16  &     51.80  & PLAT-AD (I) &     52.83  &     52.23  &     49.83  &     54.87  &     52.44  \\
\hline
                                                                                                                                  \multicolumn{ 12}{|c|}{} \\
\hline
{\bf Coded33} &     52.25  &     51.78  &     50.02  &     54.78  &     52.21  & {\bf Coded31} &     53.13  &     52.64  &     50.75  &     55.67  &     53.05  \\
\hline
  CVBF\cite{R14L} &     54.26  &     53.15  &     50.24  &     55.35  & {\bf 53.25 } &   CVBF\cite{R14L} &     55.16  &     53.76  &     50.85  &     56.06  &     53.96  \\
\hline
  ADTF\cite{R13X} &     54.07  &     53.04  &     50.29  &     55.32  &     53.18  &   ADTF\cite{R13X}&     54.80  &     53.48  &     50.94  &     55.96  &     53.80  \\
\hline
   TSF\cite{R33L} &     54.12  &     52.98  &     50.40  &     55.43  &     53.23  &        TSF\cite{R33L} &     55.00  &     53.51  &     51.02  &     56.14  &     53.92  \\
\hline
    FLAT-AD &     53.75  &     52.83  &     50.39  &     55.56  &     53.13  &     FLAT-AD &     54.72  &     53.57  &     51.02  &     56.41  &     53.93  \\
\hline
    TLAT-AD &     53.70  &     52.85  &     50.37  &     55.53  &     53.11  &     TLAT-AD &     54.70  &     53.64  &     51.02  &     56.40  &     53.94  \\
\hline
    PLAT-AD &     53.73  &     52.88  &     50.37  &     55.54  &     53.13  &     PLAT-AD &     54.75  &     53.65  &     51.01  &     56.42  &     53.96  \\
\hline
FLAT-AD (I) &     53.74  &     53.10  &     50.48  &     55.61  &     53.23  & FLAT-AD (I) &     54.63  &     53.84  &     51.06  &     56.36  & {\bf 53.97 } \\
\hline
PLAT-AD (I) &     53.75  &     53.11  &     50.47  &     55.62  & {\bf 53.24 } & PLAT-AD (I) &     54.63  &     53.84  &     51.06  &     56.36  & {\bf 53.97 } \\
\hline
\end{tabular}

\end{table*}

From Table-IV, it can be seen that the overall quality of different depth sequences filtered with our proposed method FLAT-AD (I) has the best performance, with a gain of up to 0.48 dB, while the quality of the synthesized images can be better than ADTF \cite{R13X}, but slight lower than CVBF \cite{R14L} and TSF \cite{R33L}. Meanwhile, the depth qualities of FLAT-AD, TLAT-AD, and PLAT-AD are better than ADTF, CVBF, and TSF, while TLAT-AD could better preserve boundary information than PLAT-AD and FLAT-AD. The synthesized images rendered with filtered depth maps are displayed in Fig. 3, from which we can see that the visual quality of the proposed method has superior performance compared to other methods.

The main advantage of the proposed method lies in the ability to greatly improve the quality of the depth images during the filtering than others as displayed in TABLE-I and TABLE-II. One fatal drawback of ADTF, CVBF, and TSF is that they smoothen some small but significant objects too much, and can even completely eliminate some small objects, as shown in Fig. 3 (c-e). It is obvious that the proposed method can avoid these drawbacks in Fig. 3 (f-j).

From Fig. 4, it can be seen that the CVBF spends more filtering time than ADTF, CVBF, TSF, and the proposed method, while the filtering time of the proposed FLAT-AD, PLAT-AD, FLAT-AD (I) and FLAT-AD (I) is slightly less than TSF, but more than ADTF. However, the TLAT-AD's filtering time is more than FLAT-AD, PLAT-AD, FLAT-AD (I) and PLAT-AD (I), because TLAT-AD requires to calculate the local activity in each iteration.
\begin{figure}[!t]
\centering
\includegraphics[width=3.5in]{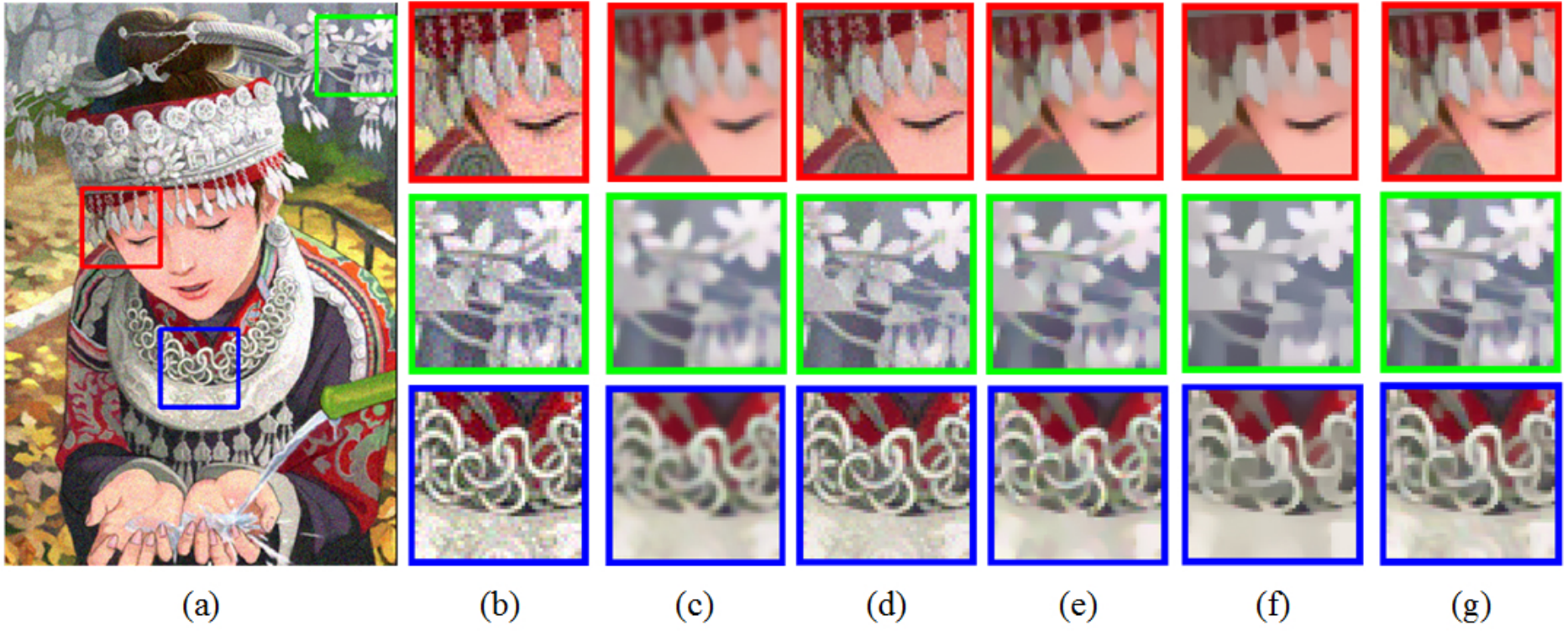}
\caption{ The visual comparison of Gaussian noise removed by several methods: (a) image containing zero mean Gaussian noise with standard deviation to be 13 from (a); (b) are enlarged from the box regions in (a); (c-g) are images filtered by respectively RBF \cite{R29K}, WBF \cite{R29K}, TV \cite{R16L}, RTV \cite{R17L}, and LAT-RTVd for (b).}
\label{Fig1}
\end{figure}

\begin{table*}[!t]
\renewcommand{\arraystretch}{1.3}
\caption{The overall performance comparison of depth images and virtual view images}
\label{table4}
\scriptsize
\centering
\begin{tabular}{|c|c|c|c|c|c|c|c|c|c|c|c|c|c|c|}
\hline
         M &                                                        \multicolumn{ 7}{|c}{Depth image} &                                            \multicolumn{ 7}{|c|}{Synthesized color image} \\
\hline
        QP &         41 &         39 &         37 &         35 &         33 &         31 & {\bf Ave.} &         41 &         39 &         37 &         35 &         33 &         31 & {\bf Ave.} \\
\hline
   Coded31 &     40.53  &     41.89  &     43.35  &     44.81  &     46.17  &     47.55  &     44.05  &     48.86  &     49.72  &     50.57  &     51.42  &     52.21  &     53.05  &     50.97  \\
\hline
  CVBF\cite{R14L} &     40.33  &     41.85  &     43.26  &     44.62  &     45.83  &     47.00  &     43.82  &     50.28  &     51.03  &     51.82  &     52.59  &     53.25  &     53.96  & {\bf 52.16 } \\
\hline
  ADTF\cite{R13X} &     40.32  &     41.68  &     43.09  &     44.44  &     45.61  &     46.68  &     43.64  &     50.14  &     51.00  &     51.78  &     52.52  &     53.18  &     53.80  &     52.07  \\
\hline
   TSF\cite{R33L} &     40.26  &     41.52  &     42.82  &     44.27  &     45.36  &     46.35  &     43.43  &     50.23  &     51.07  &     51.85  &     52.51  &     53.23  &     53.92  &     {\bf 52.14}  \\
\hline
    FLAT-AD &     40.60  &     41.85  &     43.51  &     45.20  &     46.56  &     47.91  &     44.27  &     49.98  &     50.86  &     51.67  &     52.37  &     53.13  &     53.93  &     51.99  \\
\hline
    TLAT-AD &     40.61  &     41.73  &     43.53  &     45.20  &     46.57  &     47.92  &     44.26  &     49.88  &     50.79  &     51.63  &     52.33  &     53.11  &     53.94  &     51.95  \\
\hline
    PLAT-AD &     40.65  &     42.04  &     43.56  &     45.21  &     46.57  &     47.93  &     44.33  &     49.93  &     50.82  &     51.66  &     52.35  &     53.13  &     53.96  &     51.98  \\
\hline
FLAT-AD (I) &     40.86  &     42.28  &     43.77  &     45.37  &     46.77  &     48.11  & {\bf 44.53 } &     50.40  &     50.90  &     51.81  &     52.44  &     53.23  &     53.97  &     52.13  \\
\hline
PLAT-AD (I) &     40.85  &     42.27  &     43.77  &     45.36  &     46.77  &     48.11  & { \bf 44.52}  &     50.39  &     50.89  &     51.80  &     52.44  &     53.24  &     53.97  &     52.12  \\
\hline
\end{tabular}

\end{table*}

\subsection{Clip-art compression artifact removal with LAT-AD}
LAT-AD can be used for clip-art compression artifact removal. We have tested several cartoon/clip-art images with severe compression artifacts. For clip-art compression artifact removal and image smoothness, we compare our method with TV \cite{R16L}, modified TV \cite{R3S}, and L0 gradient minimization method \cite{R18L}. Although TV could well remove the noise when the gradient along boundary is large, weak edge information is not well preserved, as shown in Fig. 5. The modified TV \cite{R3S} and the L0 gradient minimization method \cite{R18L} can preserve some weak edges, but some noise and blur still exist after being filtered by these two methods. In Fig. 5, we can see that the proposed methods are better than other methods. Our FLAT-AD (I) and FLAT-AD not only make boundaries sharper but also greatly reduce the compression artifacts, thanks to the clipped local activity tuning. The FLAT-AD (I) makes the filtered image more similar to the un-filtered image than FLAT-AD, but FLAT-AD can make edge sharper than FLAT-AD (I), which keeps the piece-wise smoothness of clip-art images.
\subsection{The denoising of contaminated depth image with LAT-RTV}

Since depth image has the properties of piece-wise smoothness and sharp discontinuity, we adopt the LAT-RTV rather than the LAT-RTVd for noise removal of depth image, because the proposed  LAT-RTV is more powerful to smoothen image than LAT-RTVd. To verify the efficiency of the proposed LAT-RTV,  ten Middlebury depth maps are tested, including: Aloe ($427\times 370$), Art ($463\times 370$), Baby1 ($413\times 370$), Baby2 ($413\times 370$), Cloth3 ($417\times 370$), Cones ($450\times 375$), Moebius ($463\times 370$), Reindeer ($447\times 370$), Teddy ($450\times 370$), and Barn1 ($432\times 370$). Here, the noise is additive white Gaussian noise, whose standard deviations are set to 4, 6, 8, 10, 15 and 20 respectively. We compare our approach with three other competing methods: nonlocal graph based transform (NLGBT) \cite{R19W}, block-matching 3D (BM3D) \cite{R20K}, and RTV \cite{R17L}, which exploit the local and nonlocal information respectively for denoising.

It has been well known that RTV has the functionality of texture removal, but as far as we know, it has never been applied into noise removal. As a matter of fact, RTV also could remove the Gaussian noise by just treating the Gaussian noise as the texture for piece-wise smoothness images. Compared with RTV, the advantages of the proposed LAT-RTV mainly come from the local activity tuning, which makes the proposed method more robust to Gaussian noise removal. It is worthy to notice that that proposed method preserves the main structure of disparity image without making boundary blur, as shown in Fig. 6 (e-f).

Table V shows the objective quality of denoising results by these methods in terms of PSNR at different noise level. The objective measure of LAT-RTV has better performance than BM3D and RTV, but has slightly lower performance than NLGBT's. As presented in Fig. 6 (c-f), we can see that our method has better edge preserving performance than others. The running time has also been tested and is reported in TABLE V, from which we can find that NLGBT has the longest filtering time compared with others, while the proposed LAT-RTV, RTV and BM3D only need several seconds.

\subsection{Image smoothing and scale-space representation with LAT-RTV}
To remove image's textures and keep structures, the proposed LAT-RTV is tuned with local activity, which can smoothen more weak edges in order to retain the main contour information. From Fig. 7, we can see that the proposed LAT-RTV can remove more textures and retain strong edges, compared with the original RTV \cite{R17L} and other four methods, including Weighted Least Squares (WLS) \cite{R21L}, Region covariance based method (RC) \cite{R24L}, Rolling   Guidance Filter (RGF) \cite{R25Q}, Robust Guided Image Filtering (RGIF) \cite{R26B}. Among these methods, RC \cite{R24L}, RGF \cite{R25Q} tend to make image's edge blurred, although they have removed many details and textures. We have also tested our LAT-RTV in three scales in Fig. 8. From this figure, we can see that proposed LAT-RTV can preserve sharp edge information and locate the edge information of main object contour, when images are represented in different scale-spaces.  Compared to RGF \cite{R25Q}, WLS \cite{R21L}, and RTV \cite{R17L}, the proposed LAT-RTV is more suitable for scale-space representation of images. Moreover, LAT-RTV has similar performance to RGIF \cite{R26B} for scale-space representation. Although both of them are achieved by optimization, they use different smoothing methods: LAT-RTV uses the features of texture and structure, and the method of RGIF considers the static and dynamic guidance's joint effects for image smoothing, so image representation in various scale-space has some diversity in the appearances, especially when some pixels have similar color information.

\begin{table*}[!t]
\renewcommand{\arraystretch}{1.3}
\caption{The objective comparison of different methods for depth image denoising and corresponding filtering time comparison}
\label{table5}
\scriptsize
\centering
\begin{tabular}{|c|c|c|c|c|c|c|c|c|c|}
\hline
  \multicolumn{ 5}{|c}{The PSNR of  filtered disparity images} &                           \multicolumn{ 5}{|c|}{Filtering time} \\
\hline
    Images/M & \multicolumn{ 1}{|c|}{BM3D \cite{R20K}} & \multicolumn{ 1}{|c|}{NLGBT \cite{R19W}} & \multicolumn{ 1}{|c|}{RTV \cite{R17L}} & \multicolumn{ 1}{|c|}{LAT-RTV} &     Images/M & \multicolumn{ 1}{|c|}{BM3D \cite{R20K}} & \multicolumn{ 1}{|c|}{NLGBT \cite{R19W}} & \multicolumn{ 1}{|c|}{RTV \cite{R17L}} & \multicolumn{ 1}{|c|}{LAT-RTV}\\
\hline
         a &      40.1  &      41.1  &      38.7  &      40.3  &          a &       1.4  &     194.9  &       1.5  &       4.5  \\
\hline
         b &      41.1  &      42.8  &      39.8  &      42.0  &          b &       1.5  &     210.9  &       1.5  &       4.5  \\
\hline
         c &      45.0  &      45.2  &      42.7  &      45.5  &          c &       1.6  &     184.7  &       1.5  &       3.3  \\
\hline
         d &      44.7  &      45.1  &      42.7  &      45.0  &          d &       1.6  &     196.0  &       1.5  &       3.9  \\
\hline
         e &      44.8  &      45.0  &      41.4  &      44.8  &          e &       2.0  &     260.0  &       1.7  &       4.1  \\
\hline
         f &      42.7  &      43.8  &      39.1  &      42.3  &          f &       1.9  &     262.7  &       1.6  &       5.0  \\
\hline
         g &      43.4  &      43.5  &      40.4  &      43.1  &          g &       2.2  &     309.0  &       1.8  &       5.2  \\
\hline
         h &      43.3  &      44.1  &      40.5  &      43.1  &          h &       2.0  &     272.5  &       1.9  &       5.6  \\
\hline
         i &      42.7  &      42.9  &      39.3  &      42.1  &          i &       2.0  &     287.7  &       1.9  &       5.5  \\
\hline
         j &      47.1  &      46.9  &      45.5  &      47.7  &          j &       2.1  &     259.8  &       1.7  &       3.7  \\
\hline
      mean &      43.5  & {\bf 44.0 } &      41.0  &      43.6  &       mean &       1.8  &     243.8  & {\bf 1.7 } &       4.5  \\
\hline
\end{tabular}

\end{table*}
\begin{table*}[!t]
\renewcommand{\arraystretch}{1.3}
\caption{The objective comparison of noisy image filtered by different methods}
\label{table6}
\scriptsize
\centering
\begin{tabular}{|c|c|c|c|c|c|c|c|c|c|c|c|c|c|}
\hline
\multicolumn{ 7}{|c}{Standard deviation=13} & \multicolumn{ 7}{|c}{Standard deviation=26} \\
\hline
\multicolumn{ 1}{|c|}{Image} & \multicolumn{ 1}{|c|}{Noisy} &        \multicolumn{ 1}{|c|}{RBF \cite{R29K}} &        \multicolumn{ 1}{|c|}{WBF \cite{R29K}} &         \multicolumn{ 1}{|c|}{TV \cite{R16L}} &       \multicolumn{ 1}{|c|}{RTV \cite{R17L}} & \multicolumn{ 1}{|c|}{LAT-RTVd} & \multicolumn{ 1}{|c|}{Image} & \multicolumn{ 1}{|c|}{Noisy} &        \multicolumn{ 1}{|c|}{RBF \cite{R29K}} &        \multicolumn{ 1}{|c|}{WBF \cite{R29K}} &         \multicolumn{ 1}{|c|}{TV \cite {R16L}} &        \multicolumn{ 1}{|c|}{RTV \cite{R17L}} & \multicolumn{ 1}{|c|}{LAT-RTVd} \\
\hline
       (a) &      26.04 &      31.75 &      32.87 &      29.15 &      31.55 &      33.51 &        (a) &      20.14 &      30.36 &      30.45 &      27.66 &      28.31 &      30.19 \\
\hline
       (b) &      26.07 &      25.71 &      29.48 &      25.46 &      24.61 &      27.80 &        (b) &      20.21 &      25.25 &      26.17 &      24.78 &      23.87 &      25.69 \\
\hline
       (c) &      26.17 &      30.27 &      31.71 &      29.39 &      30.19 &      32.02 &        (c) &      20.31 &      29.15 &      29.27 &      27.74 &      28.45 &      29.09 \\
\hline
       (d) &      26.07 &       30.4 &      31.65 &      30.23 &      29.62 &      31.97 &        (d) &      20.22 &      29.49 &      29.60 &      28.44 &       28.30 &      29.22 \\
\hline
       (e) &      26.36 &      26.67 &      29.54 &      26.55 &      26.68 &      29.66 &        (e) &      20.49 &      25.92 &      26.53 &      25.50 &      25.31 &      26.94 \\
\hline
       (f) &      26.17 &      24.39 &      28.44 &       23.08 &      24.78 &      27.92 &        (f) &      20.36 &      23.74 &      24.92 &      22.61 &      22.73 &      25.06 \\
\hline
       (g) &      26.26 &      27.79 &      29.27 &       26.62 &      26.77 &      30.65 &        (g) &      20.31 &      26.82 &      27.13 &      25.48 &      24.93 &      27.26 \\
\hline
       (h) &      26.18 &      27.72 &      30.17 &      26.70 &      27.01 &      30.62 &        (h) &      20.44 &      26.85 &      27.29 &      25.71 &      25.18 &      27.46 \\
\hline
       (i) &      26.62 &      30.85 &      31.52 &      31.07 &      29.31 &       33.10 &        (i) &      20.86 &      28.72 &      28.14 &      28.44 &      26.41 &      28.78 \\
\hline
       (j) &      26.27 &      30.58 &      31.56 &      29.65 &      30.21 &      32.95 &        (j) &      20.32 &      29.17 &      29.24 &      27.78 &      27.94 &      29.13 \\
\hline
      Ave. &      26.22 &      28.61 &      30.62 &      27.79 &      28.07 &      {\bf 31.02} &       Ave. &      20.37 &      27.55 &      27.87 &      26.41 &      26.14 &      {\bf 27.88} \\
\hline
\end{tabular}

\end{table*}

\subsection{Image denoising with LAT-RTVd}

Ten images are used to test image denoising, including: Monarch, Barbara, Pepper, Lena, Man, Comic, Zebra, Flowers, Bird, Boats. The noise is zero mean Gaussian noise with standard deviation of 13 and 26. We compare the proposed approach with four other methods. The non-linear combination of the local activity and gradient information in the LAT-RTVd catch the location of the noise, so Gaussian noises can be removed and fine details are still retained, but RTV only tends to smooth the texture to preserve image’s structure. This is shown in Fig. 9, where three other methods including RBF \cite{R29K}, WBF \cite{R29K}, and TV \cite {R16L}, are also compared with the proposed LAT-RTVd. From Fig. 9, and TABLE VI, we can see that both objective quality and visual quality of the proposed method for denoising have better performance than other methods and the total gains of noisy image's PSNR can be up to 7.51 dB compared with noisy image.

\section{Conclusion}
In this paper, two local activity-tuned frameworks are introduced. First, a robust local activity-tuned anisotropic diffusion is proposed to control the diffusion for depth artifact's removal. Secondly, our local activity-tuned relative total variation framework achieves good performance for image smoothing and represents the image in different scale-space and it has been used for depth image denoising. From these applications, we can see that proposed LAT-AD, LAT-RTV and LAT-RTVd has good performance for image smoothing and noise removal. The local activity-tuned strategy can be applied into other schemes, which will be explored in our future works.




\bibliographystyle{IEEEtran}
\bibliography{IEEEfull,VTIFbibfile}
\end{document}